\documentclass[letterpaper, 10 pt, conference]{ieeeconf}  

\IEEEoverridecommandlockouts                              

\usepackage{graphicx} 
\usepackage{amsmath} 
\usepackage{amssymb}  
\usepackage{xcolor}
\usepackage{booktabs}
\usepackage{caption}
\usepackage{subcaption}
\usepackage{hyperref}

\newcommand{\yf}[1]{\textcolor{purple}{[YF:#1]}}

\newcommand{\hide}[1]{}

\definecolor{darkorange}{RGB}{255,102,153}
\definecolor{darkgreen}{RGB}{102,204,51}
\definecolor{lightblue}{RGB}{51,153,255}
\definecolor{darkpurple}{RGB}{204, 153, 255}

\title{\LARGE \bf
Long-horizon Locomotion and Manipulation on a Quadrupedal Robot with Large Language Models
}

\author{Yutao Ouyang$^{1*}$, Jinhan Li$^{1*}$, Yunfei Li$^{1}$, Zhongyu Li$^{3}$, Chao Yu$^{1}$, Koushil Sreenath$^{3}$, Yi Wu$^{1,2, \dag}$
\thanks{*Equal contribution}
\thanks{\dag Corresponding author jxwuyi@mail.tsinghua.edu.cn}
\thanks{$^{1}$Tsinghua University, Beijing, China
        {\tt\small }}%
\thanks{$^{2}$Shanghai Qi Zhi Institute, Shanghai, China
        {\tt\small }}%
\thanks{$^{3}$University of California, Berkeley, CA, USA}
}

\begin{document}

\maketitle
\thispagestyle{empty}
\pagestyle{empty}
\begin{abstract}

We present a large language model (LLM) based system to empower quadrupedal robots with problem-solving abilities for long-horizon tasks beyond short-term motions. Long-horizon tasks for quadrupeds are challenging since they require both a high-level understanding of the semantics of the problem for task planning and a broad range of locomotion and manipulation skills to interact with the environment. Our system builds a high-level reasoning layer with large language models, which generates hybrid discrete-continuous plans as robot code from task descriptions. It comprises multiple LLM agents: a semantic planner that sketches a plan, a parameter calculator that predicts arguments in the plan, a code generator that converts the plan into executable robot code, and a replanner that handles execution failures or human interventions. At the low level, we adopt reinforcement learning to train a set of motion planning and control skills to unleash the flexibility of quadrupeds for rich environment interactions. Our system is tested on long-horizon tasks that are infeasible to complete with one single skill. Simulation and real-world experiments show that it successfully figures out multi-step strategies and demonstrates non-trivial behaviors, including building tools or notifying a human for help. Demos are available on our project page: \href{https://sites.google.com/view/long-horizon-robot}{https://sites.google.com/view/long-horizon-robot}.

\end{abstract}

\section{Introduction}
Quadrupedal robots have demonstrated advantages in mobility and versatility using a variety of skills.
The field has witnessed exciting advancement in developing locomotion controllers to enhance traversability~\cite{de2006quadrupedal,fankhauser2018robust,lee2020learning} and manipulation controllers to facilitate physical interactions with the world~\cite{shi2021circus,sombolestan2023hierarchical,arm2024pedipulate}.   
Although much progress has been made in the acquisition of specific motion skills, 
we would like to push the autonomy of the robot to a higher level beyond the individual skills and ask: 
How can a quadruped robot combine its locomotion and manipulation abilities to tackle challenging problems necessitating long-horizon planning and strategic interactions with the environment?

An example of a long-horizon task may be turning the lights off before exiting an office, a scenario demanding a strategic combination of actions such as climbing to reach a button, pressing an object, and walking. Such a long-term maneuver poses significant challenges both in high-level reasoning and in low-level motion planning and control. 
\hide{Consider a button-pressing task illustrated in Fig.~\ref{fig:overview}. The quadruped robot aims to press a button on the wall to open the door or to turn on/off the light, but the height of the button is beyond the reaching ability of the robot. To complete the task, the robot should build some stairs from the boxes in the room, then climb to the top of the boxes, and finally pitch up to press the button with its leg.}

For high-level reasoning, the agent must devise a multi-step strategy that takes into account both physical affordances and skill limitations, all without the aid of immediate learning signals,
thus encountering substantial exploration hurdles in long-term problem-solving.
Furthermore, the vast decision space of the high-level policy exacerbates the challenges. 
The policy needs to simultaneously determine which skill to execute (e.g., walking or pitching up) and how to execute it (e.g., specifying continuous parameters for the target position).
Prior research often relies on demonstrations~\cite{cheng23legs} or heuristics in search~\cite{garrett2015ffrob} to reduce the computation burden.
As for low-level control, the quadrupeds' nonlinearity and high dimensionality pose challenges in acquiring versatile behaviors necessary to enable complex environmental interactions.

\begin{figure}
    \centering
    \includegraphics[width=0.9\linewidth]{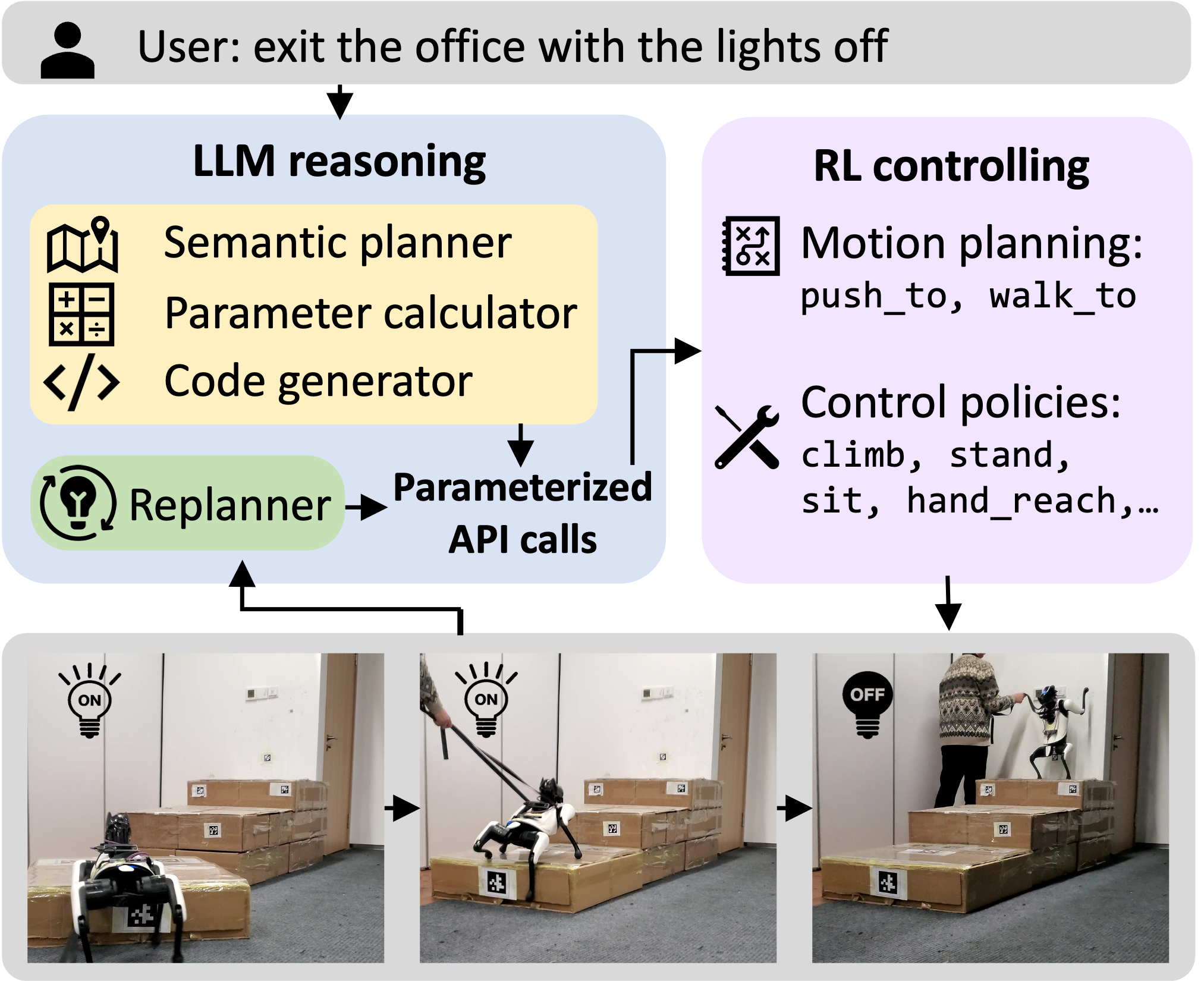}
    \captionsetup{font={small}}
    \caption{Overview of the hierarchical system for long-horizon loco-manipulation task. The system is built up from a \textcolor[rgb]{0.182, 0.333, 0.591}{reasoning layer} for task planning\hide{decomposition}\hide{ (yellow)} and a \textcolor[rgb]{0.439,0.188,0.627}{controlling layer} for skill execution\hide{ (purple)}. Given the language description of the long-horizon task\hide{ (top)}, \hide{a cascade of} \textcolor[rgb]{1, 0.753, 0.}{three cascaded LLM agents} perform high-level task decomposition and generate a complete plan. A \textcolor[rgb]{0.329, 0.510, 0.208}{replanner} is introduced to handle unexpected situations.\hide{planning and produce function calls of parameterized robot skills. }
    \hide{It is composed of a semantic planner that proposes a solution consisting of \textcolor{darkorange}{branches} conditioned on environment specifications and \textcolor{darkgreen}{primitive actions}, a parameter calculator that fills in \textcolor{lightblue}{arguments} for the actions, a code generator to summarize the plan as executable code.}
    The reasoning layer generates function calls for parameterized robot skills, and the controlling layer
     instantiates the mid-level motion planning and low-level controlling skills with RL. 
    }
    \vspace{-1.2em}
    \label{fig:intro:overview}
\end{figure}

We propose a hierarchical system to address the challenges in long-horizon loco-manipulation problems for quadrupeds as illustrated in Fig.~\ref{fig:intro:overview}. 
At the high level, we harness the planning ability of pretrained large language models (LLMs) for abstract reasoning over the long horizon.
Leveraging the wealth of encoded common-sense knowledge, the LLM-based reasoning layer can directly predict high-level strategies without demonstrations or searching heuristics.
To better ground to robotic problems, we design the reasoning module as a cascade of multiple LLM agents. 
A semantic planner sketches a discrete plan, while a parameter calculator predicts continuous parameters.
Subsequently, a code generator encapsulates the plan into function calls for robot skills. 
To achieve high-level closed-loop planning, we introduce a replanner to handle unexpected events.
At the low level, we train a spectrum of motion planning and control policies with model-free reinforcement learning, spanning from quadrupedal locomotion to bipedal manipulation.

Our contributions are as follows:
We introduce a new closed-loop reasoning architecture with multiple LLM agents, which focuses on long-sequence robotic problems. It is capable of generating comprehensive plans adaptable to diverse scenes and dynamically adjusting through feedback.
Our system chains both locomotion and manipulation skills, and enables more complex tasks that require nuanced physical interaction.
With the high-level reasoning capabilities of LLM, it can address complex long-horizon tasks involving more than ten function calls of skills.
We extensively validate our design components in four novel and challenging long-horizon tasks for quadrupedal robots including manipulating a light switch at a height, package delivery, bridge building, and taking an elevator. 
Our method outperforms all the baselines, demonstrating a success rate of over 70\% in simulation and successful deployment in the real world.

\section{Related Work}
 
\textbf{Quadrupedal locomotion and manipulation:} 
Developing versatile motions over quadrupedal robots to make them work like real animals has attracted much research interest.
A lot of work focuses on robust and agile locomotion skills that can walk on challenging terrains~\cite{hwangbo2019learning,lee2020learning,miki2022learning} and perform extreme parkour~\cite{zhuang2023robot,cheng2023extreme}. 
Some recent works develop manipulation skills for quadrupeds to better interact with the environment,
such as learning manipulation via locomotion to push heavy objects~\cite{jeon2023learning}, 
manipulating with one or two legs while balancing on other legs~\cite{schwarke2023curiosity,cheng23legs,arm2024pedipulate},
or manipulating with a gripper mounted on the robot~\cite{bellicoso2019alma,fu2023deep}.   
There are also works that learn mid-level motion planning policies over these low-level motion controllers for more complex skills, such as soccer manipulation~\cite{ji2022hierarchical,huang2023creating}. We focus on empowering quadrupeds with long-horizon problem-solving abilities beyond motion planning and control skills for specific tasks. The problems considered in this work require an abstract task-planning ability that reasons over the semantics. The desired strategy, such as building stairs alongside the wall so that the robot can climb up and reach the light switch, requires common-sense knowledge about the physical constraints of the environment and the robot's capabilities and is challenging to discover from scratch.

\textbf{Tackling long-horizon challenge in robotic tasks:} 
Traditionally, Task and Motion Planning (TAMP)~\cite{garrett2021integrated} methods are adopted to get plans for long-horizon tasks that involve a hybrid of discrete tasks and continuous parameters, but typically require entire environment states and suffer from heavy computational burden~\cite{hartmann2022long}.
Training hierarchical policies incorporating temporal abstractions is another approach to tackle long-horizon problems~\cite{gupta2020relay,yang2021hierarchical,zhu2022bottom}.
Some other works adopt behavior trees~\cite{cheng23legs} or skill chaining~\cite{lee2022adversarial} to address the long-horizon challenge.

Recently, pretrained large language models~\cite{brown2020language,ouyang2022training,chowdhery2023palm} have demonstrated impressive reasoning abilities to plan over robot skills for solving long-horizon tasks. A line of work uses LLMs to decompose long-range tasks into step-by-step plans in natural language~\cite{brohan2023can,huang2022language,huang2023inner}, and grounds the language plan to robot execution with language-conditioned control policies. With the development of code-completion LLMs, researchers have also considered using Pythonic code as the interface between LLM reasoning and robot skills~\cite{zeng2022socratic,liang2023code} since it can be more directly grounded to available robot APIs and allows more fine-grained behaviors using numerical arguments in function calls.
We similarly leverage LLMs to generate robot code for completing long-horizon tasks that involve physical constraints in this work and separate semantic planning, numerical argument computing, and code writing to different LLM agents.

The most relevant work to us is RoboTool~\cite{xu2023creative} which also reasons over long-range physical puzzles with multiple LLMs. 
However, RoboTool’s open-loop planning produces fixed plans for specific scene configurations, limiting its adaptability to complex tasks and unexpected events. In contrast, our approach generates branching conditions within the same code to handle diverse configurations and incorporates a replanner that dynamically adjusts the plan based on execution feedback.
Additionally, RoboTool only leverages the basic quadrupedal locomotion skills, which essentially treats the quadruped robot as a mobile platform and does not fully utilize its high dimensionality. 
\hide{Furthermore, RoboTool is designed to generate plans specific to certain scene configurations, while our method can generate branching conditions to handle various scene configurations with the same piece of code.} 
As we will see in Sec.~\ref{sec:Ablation Study Analysis}, our method shows 
more robust planning performance.


\section{Method}
\begin{figure*}
    \centering
    \includegraphics[width=\textwidth]{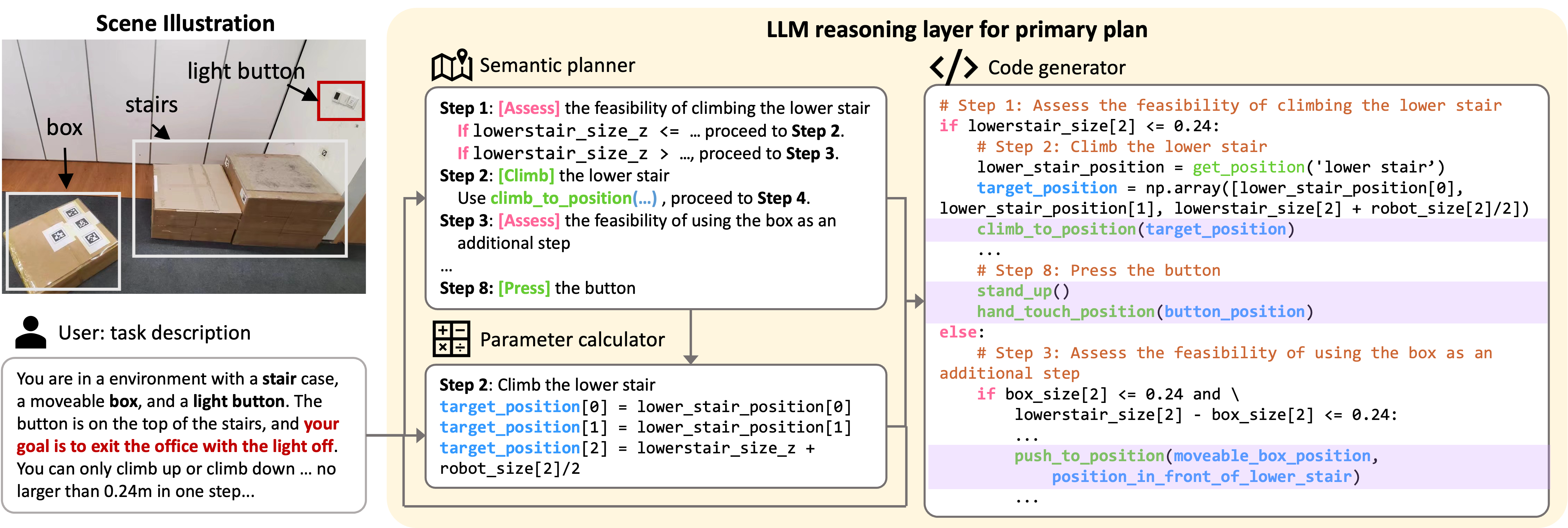}
    \captionsetup{font={small}}
    \caption{Illustration of the LLM-based high-level reasoning layer that generates primary hybrid discrete-continuous plans from the task description in language. It is composed of a semantic planner that proposes a solution consisting of \textcolor{darkorange}{branches} conditioned on environment specifications and \textcolor{darkgreen}{primitive actions}, a parameter calculator that fills in \textcolor{lightblue}{arguments} for the actions, a code generator to summarize the plan as executable robot code. The texts are abbreviated generated contents from the LLMs.
    }
    \label{fig:overview}
    \vspace{-1.5em}
\end{figure*}

In this work, we develop a hierarchical system on a quadruped robot to solve challenging long-horizon tasks that require a smart combination of dynamic skills\hide{, such as opening a door with the exit button high on the wall}. For high-level reasoning over the long horizon\hide{, physical affordance and controller capabilities,}, we design a cascade of LLMs to figure out hybrid discrete-continuous plans represented as parameterized robot skill API calls. \hide{The abstract plan sketching and the accurate parameter calculation are assigned to separate LLM modules to reduce the reasoning burden when generating hybrid discrete and continuous plans. }
At the low level, we instantiate motion planning and control for locomotion and manipulation skills with model-free RL. \hide{An overview of our system is shown in Fig.~\ref{fig:overview}.}

\subsection{High-level reasoning for long-horizon tasks with LLMs}
To address the challenging reasoning and grounding problem, we design four LLM agents with different focuses as the high-level reasoning module. The first three cascaded agents, semantic planner, parameter calculator, and code generator, work together to generate a complete initial plan as shown in Fig.~\ref{fig:overview}. The semantic planner focuses on task decomposition with constraints from physical affordances and robot capabilities and sketches a plan of primitive skills for the long-horizon problem.
The parameter calculator figures out the precise arguments for the primitives in the plan. The code generator converts the plan and the arguments into one single piece of executable code that can strategically combine the available low-level skills to handle the long-horizon problem.
In addition, a replanner is introduced as the fourth agent to enable closed-loop control, addressing unexpected situations by 
adjusting the plan and recalculating necessary parameters to generate a new code plan dynamically during execution

We use GPT-4-turbo-preview~\cite{achiam2023gpt} for all the LLM agents\hide{\footnote{We also tried other LLMs such as GPT-3.5~\cite{ouyang2022training} and Kimichat~\cite{kimichat}. GPT-3.5 struggles to generate feasible abstract plans due to its deficiency in reasoning. Kimichat can generate reasonable plans given the environment descriptions and constraints but fails to calculate the accurate parameters even when prompted with detailed instructions.\yf{can be removed}
}}.

\subsubsection{Semantic planner}
The agent reasons at an abstract level to predict a feasible plan from natural language descriptions of the task setup and the available robot skills. The required strategy may differ according to variations in the environment configurations, e.g., the robot can directly climb up stairs if the slope is within its ability but it needs to seek a box to step on before climbing otherwise. Different from previous approaches that run planning for specific environment configurations, we ask the planner to consider different conditions and use ``if-else'' clauses to unify all strategies into a single plan. Therefore, our generated plan could merit generalization over different setups at test time.

We first provide the planner with descriptions of each motion skill, the physical rules to follow, and an example showing the output format. We then give environment descriptions, specifically the information about the objects contained in the environment, the task to complete, and the limit of the robot's low-level skills. Note that the environment information does not contain numerical values about the objects. Instead, we encourage the planner to use ``if-else'' clauses in the plan to consider different cases.

The planner formulates the plan as a workflow with branches and jumps. Each step is either a feasibility check or an action step. A check step redirects the control flow to different steps depending on the environment configuration or raises an error if the task is deemed unsolvable given all the constraints. An action step invokes one or more parameterized skills to be executed sequentially, but the corresponding parameters may be incorrect or undefined, e.g., the predicted target position for the climbing skill in step 2 does not consider the size of the robot. The parameters are handled by a calculator described next.

\subsubsection{Parameter calculator}
We leverage a second LLM to obtain the correct parameters for the invoked robot skills in the plan. Previous research~\cite{liang2023code,yu2023language,xu2023creative} find that a single LLM is not proficient at reasoning across multiple levels of abstraction. We similarly observed that the planner for task decomposition is erroneous regarding detailed parameters for the skills, therefore assigning the calculation to another LLM.

The parameter calculator is designed to generate formulas that can be executed to calculate the 3D targets from variables given in the environment description. It takes the predicted plan along with the original environment description as input, and is prompted with rules for computation and one example output. For each action step, the calculator first parses out the primitive skill that needs to compute arguments, then thinks about how to do the calculation verbally, and finally gives the formulas for every dimension of the arguments. The snapshot of its output can be seen from Fig.~\ref{fig:overview}.

\subsubsection{Code generator} 
Given the abstract plan and the detailed parameters for the skills in the plan, we use a third LLM agent to convert them into executable Python code that can directly orchestrate the low-level locomotion and manipulation skills.
We prompt the code generator with documented function definitions and global variables that can be used in the code, rules for code writing, and one dummy example for formatting. The available functions are the skill set of the robot and a perception function to query the position of an object. The global variables are geometrical properties of objects, whose values will be assigned at test time. We then provide the abstract plan, the parameters, and the environment description to the code generator. 
The code output is shown in Fig.~\ref{fig:overview}. The code generator interprets the control flow in the plan as conditioning blocks, triggers relevant skills in each branch, and performs appropriate arithmetic operations to obtain parameters.

\subsubsection{Replanner}
We implement a replanner to handle unexpected events during execution, such as human interruptions that change the task goal or skill execution failures. Open-loop planning lacks the ability to adapt to execution feedback, making it vulnerable to systemic errors and unforeseen events. To mitigate these, when an unexpected event occurs, the robot first invokes a recovery strategy and gathers relevant error information, including the spatial information about the robot and the associated object of the failed skill. In the case of a human interruption, the robot receives new language prompt instructions to adapt its behavior accordingly. The replanner then leverages the original environment description and the previously generated primary code plan to produce a new plan adapted to the current situation. By extracting necessary knowledge from the primary code plan, it bypasses the first three cascaded structure, reducing the overhead of plan adjustments. By analyzing execution errors, the replanner determines whether to refine the plan or recalibrate parameters. This closed-loop approach allows the robot to iteratively correct its actions, enhancing adaptability and ensuring robust task execution.

\subsection{Low-level motion planning and control with RL}
We implement robot skill APIs used in the high-level reasoning layer with RL-based motion planning and control policies. 
We first train control policies for short-term motions including quadrupedal locomotion on terrains, i.e., \texttt{climb\_to\_position}, 
and bipedal locomotion and manipulation skills \texttt{stand\_up}, \texttt{hand\_touch\_position, \texttt{sit\_down}}. Building upon the control policies, we develop motion planning policies for mid-term strategies such as \texttt{push\_to\_position} for moving objects and \texttt{walk\_to\_position} for navigation with collision avoidance. 
To switch between the skills more robustly, we further fine-tune the policies by initializing each of them from the termination states of the preceding skill following the predicted order in the LLM-generated code.
All these skills are trained with PPO~\cite{schulman2017proximal} in domain randomized simulation. 

\subsubsection{Learning control policies for short-term motions}

\textbf{Quadrupedal locomotion} 
is formulated as a velocity-tracking motion, and is trained with a multi-stage curriculum of increasingly challenging terrains to efficiently learn a robust policy. 
In the first stage, a walking policy is trained on a flat surface following the same reward specification of \cite{margolis2023walk}. It is then transferred to the second stage to learn locomotion over static stairs where the stair height ranges from 0 to 0.35m. 
After the policy is capable of climbing up the highest stairs, it is transferred and continues to be trained in the final stage where the stairs are built from multiple movable boxes. The policy is trained to climb across the boxes while avoiding moving the boxes at the same time. This final stage encourages the robot to learn a gentle and robust climbing strategy without stepping on the edges of the terrain.

\textbf{Bipedal locomotion and manipulation}
To facilitate interaction with objects, we develop bipedal locomotion and manipulation policies that control the quadrupedal robot to reach target positions with its front legs while standing on its hind legs. 
We first train a bipedal locomotion policy that can pitch up from a four-leg standing pose, then track a linear bipedal walking velocity using the method in~\cite{li2023learning}.
Afterward, we train a bipedal manipulation policy to reach target positions using one front leg with a summation of three categories of reward terms. The first category is position-tracking rewards that encourage the robot to move its left front toe to the desired position as close as possible. To reduce jitter around the desired position, the robot is also given a bonus reward linear to the number of consecutive steps during which the distance to the desired position is within a threshold. The second category shapes the robot base and the other three legs into an upright standing pose. 
The third category is regularization terms that penalize abrupt actions, large joint velocities, and dangerous collisions.
The policy is trained from states in bipedal standing poses with the robot base pose and the joint positions sampled from a hand-crafted range.  
To transit from bipedal to quadrupedal standing poses, we also train a sit-down policy following~\cite{li2023learning}.

\begin{figure}[t]
    \centering
    \includegraphics[width=\linewidth]{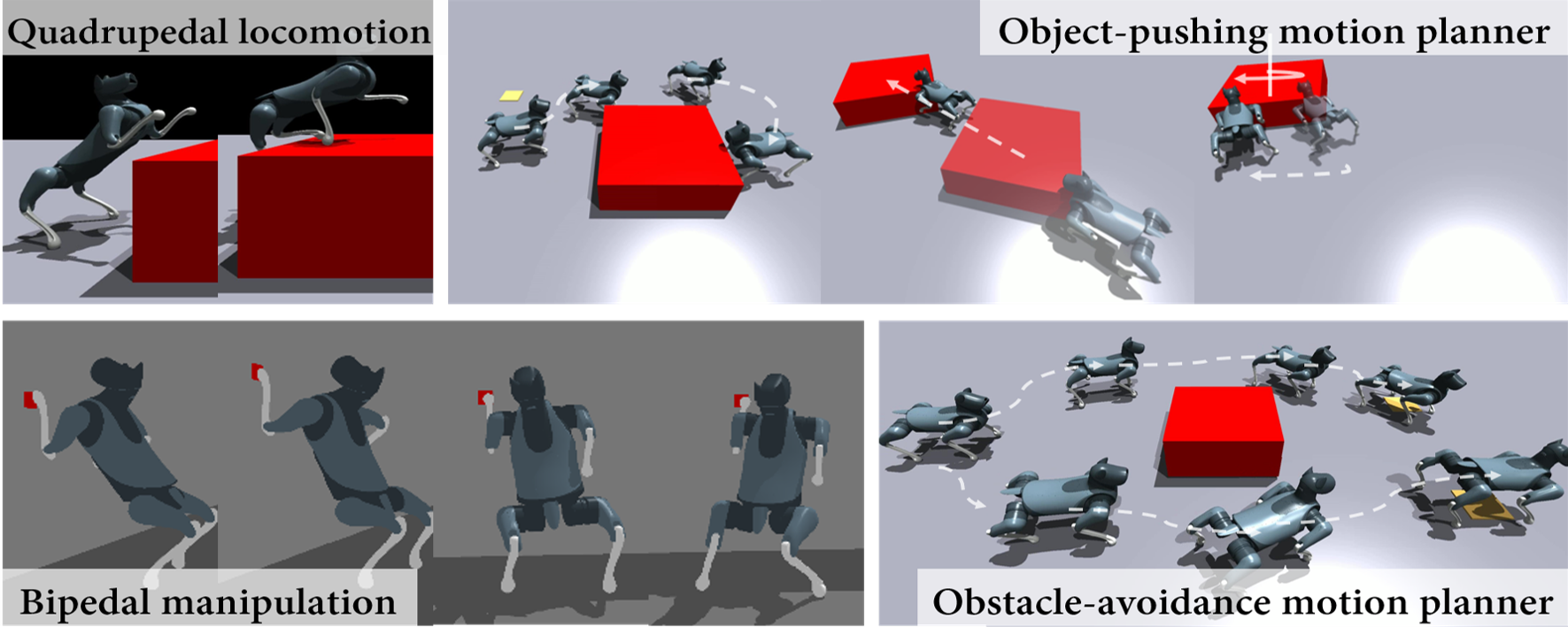}
    \captionsetup{font={small}}
    \caption{Versatile behaviors driven by the RL skill repertoire.}
    \label{fig:expr:single_skill}
    \vspace{-1.5em}
\end{figure}

\subsubsection{Learning motion planning policies with hierarchical RL}
We build object-pushing and obstacle-avoiding skills with hierarchical RL, both trained as mid-level motion planning policies over the previously obtained quadrupedal locomotion controllers. The \textbf{object-pushing policy} aims to manipulate an object to a desired pose. It is trained to predict the linear and angular velocity for the locomotion controller every 0.5 seconds. The policy takes low-dimensional states including the current and the desired object pose, the robot body pose and the object size as input. The primary reward encourages a shorter distance between the current and the desired object pose. We additionally credit the robot to face towards the object so that it can keep the vision of the object.  

As for the \textbf{obstacle-avoiding policy}, it aims to control the robot to move to a desired position and orientation while avoiding collision with an obstacle in the scene. The policy shares the same action space as the object-pushing policy and predicts actions every 0.2 seconds. The observation similarly contains the current pose of the obstacle, the current and the desired pose of the robot base and the size of the obstacle. The rewards consist of a pose-tracking term that encourages the robot to match the target pose and an obstacle-avoidance term that penalizes the robot for being too close to the object.  

\begin{figure*}[t]
    \centering
    \includegraphics[width=\textwidth]{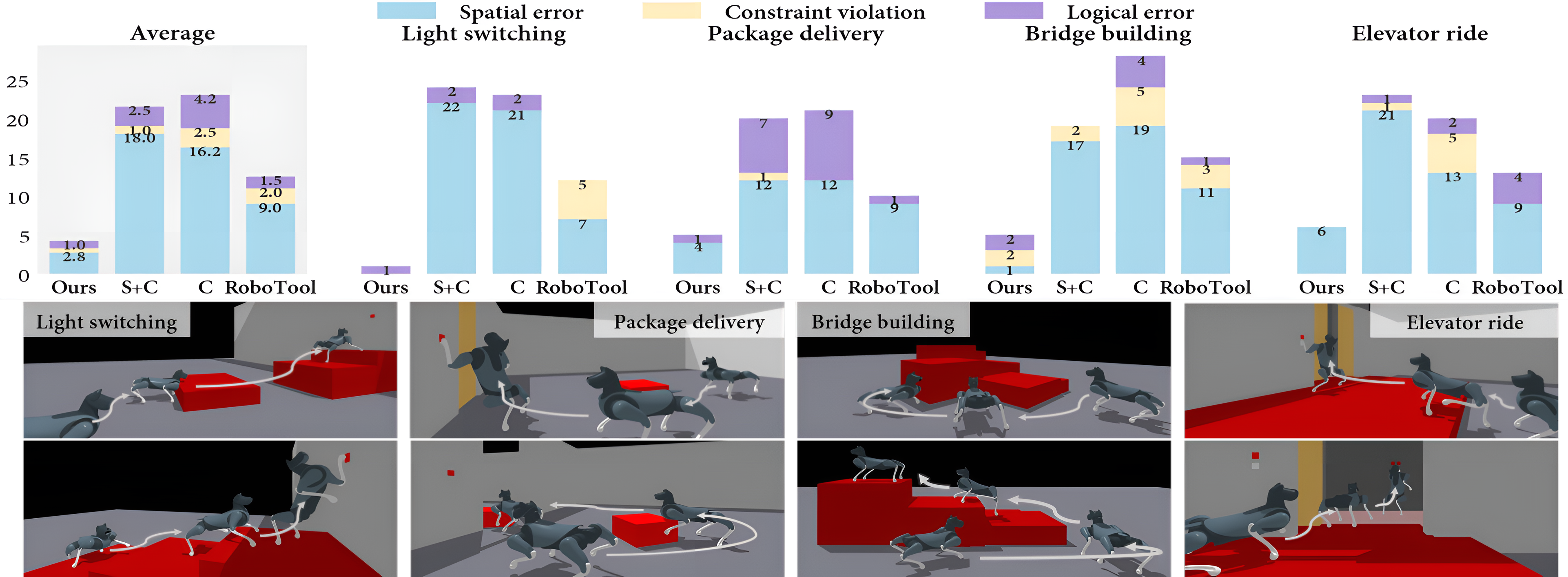}
    \captionsetup{font={small}}
    \caption{Total number of occurrences for different error types in generated primary plan code, with ten samples per variant. \textcolor[rgb]{0.384,0.559,0.547}{Spatial error} refers to errors in perceiving spatial relations leading to miscalculations, \textcolor[rgb]{1, 0.753, 0.}{constraint violation} refers to ignoring the given constraints, and \textcolor[rgb]{0.439,0.188,0.627}{logical error} refers to faulty logical reasoning. The lower part shows the execution of the 4 tasks in simulation.
    }
    \label{fig:expr:llm_exp_and_sim}
    \vspace{-1.5em} 
\end{figure*}

\subsubsection{Chaining RL skills and recovery} Directly executing different skills sequentially often leads to suboptimal behavior since the terminal state distribution of one policy may not perfectly match the initial state distribution of the next policy.
For instance, the preceding box-pushing policy often terminates with the box deviating from the ideal pose right beneath the existing stairs. Executing the climbing skill from the out-of-distribution states will lead to missteps and falls, thus resulting in lower success rates as more skills are executed. 
To mitigate the impact of the preceding strategy’s termination state on the subsequent strategy, we collect a set of termination states of the preceding policy of each policy, then fine-tune the policy from a mixture of the original initial state distribution and the collected states. Note that we only run fine-tuning for the pair of policies that will be executed sequentially according to the plan generated by the reasoning layer to reduce the computation burden.

We also train an RL-based recovery policy to restore a stable quadrupedal stance. By sampling states during execution, especially in risky skills like climbing and hand touching, we define diverse initial conditions. Rewards then guide smooth transitions to target joint states, enabling rapid failure recovery and ensuring readiness for replanning.

\section{Experiments}
\subsection{Simulation experiment setting}

\subsubsection{Baselines}
We first verify the proposed design of the high-level reasoning layer composed of a cascade of LLM agents in the simulation. 
We present our method \textbf{LLM (S+P+C+R)}, which incorporates a semantic planner, a parameter calculator, a code generator, and a replanner. 
We impose a time limit for task completion, within which the replanner can be called if a skill execution fails, and then compare it with the following variants.

\begin{itemize}
    \item \textbf{LLM (S+P+C)} does not allow the invocation of replanner but retains a full primary plan generation structure.
    \item \textbf{LLM (S+C)} that removes the parameter calculator agent, i.e., passing the output of the planner and the task description to the coder. The coder needs to calculate the parameters to write functional programs;
    \item \textbf{LLM (C)} with code generator only, i.e., directly generating code from the task description.
    \item \textbf{RoboTool~\cite{xu2023creative}} which is an LLM-based planning framework focusing on tool-use tasks. 
    \item \textbf{Hierarchical RL~\cite{huang2023creating}:} We train a high-level RL policy that jointly predicts the discrete low-level skill ID and its continuous parameters every 0.1s.

\end{itemize}

To ensure a fair comparison of high-level reasoning with RoboTool, we constructed LLM agents following RoboTool’s structure and prompts, provided RoboTool with the same low-level skills and supplemented their original rules with additional information and computation rules from our method. We provided specific numerical parameters according to their environment description setting, which are not required by our LLM reasoning layer. For hierarchical RL,  We implement a two-head MLP with separate output layers for discrete and continuous predictions respectively. The policy is optimized using PPO~\cite{schulman2017proximal} and trained with a dense reward based on the aforementioned distance metric.  

\begin{table*}[h]
\captionsetup{font={small}}
\setlength{\abovecaptionskip}{-0.05cm}
\setlength{\belowcaptionskip}{+0.05cm}
\caption{Ablation studies on the reasoning layer. The mean and standard deviation of the success rate and a distance metric in four tasks across 3 seeds are reported. The composition of multiple LLM agents is critical to the good performance of the system. LLM-based 
reasoning performs better than an RL-based high-level policy.  
}
\label{tab:llm-ablation}
\begin{center}
\begin{tabular}{lcccccccc}
\toprule
 & \multicolumn{2}{c}{Light switching} & \multicolumn{2}{c}{Package delivery} & \multicolumn{2}{c}{Bridge building} & \multicolumn{2}{c}{Elevator ride}\\
 & Success $\uparrow$ & NormDist $\downarrow$ & Success $\uparrow$ & NormDist$\downarrow$ & Success $\uparrow$ & NormDist $\downarrow$ & Success $\uparrow$ & NormDist$\downarrow$ \\
\midrule
LLM (S+P+C+R) & \textbf{0.877$\pm$0.015} &  \textbf{0.024$\pm$0.003}& \textbf{0.823$\pm$0.035} & 0.204$\pm$0.049 & \textbf{0.770$\pm$0.052} &  0.100$\pm$0.021 & \textbf{0.580$\pm$0.020} & \textbf{0.136$\pm$0.008} \\
LLM (S+P+C) & 0.747$\pm$0.012 &  0.032$\pm$0.004& 0.740$\pm$0.022 & \textbf{0.119$\pm$0.007} & 0.673$\pm$0.047 &  \textbf{0.032$\pm$0.004}& 0.530$\pm$0.042 & 0.159$\pm$0.019 \\
LLM (S+C) & 0.0$\pm$0.0 & 0.326$\pm$0.002 & 0.0$\pm$0.0 & 0.697$\pm$0.408 & 0.467$\pm$0.316 & 0.231$\pm$0.141 & 0.0$\pm$0.0 & 0.505$\pm$0.007\\
LLM (C) & 0.0$\pm$0.0 & 0.397$\pm$0.083 & 0.0$\pm$0.0 & 0.483$\pm$0.249  
& 0.0$\pm$0.0 & 0.496$\pm$0.044 & 0.0$\pm$0.0 & 0.612$\pm$0.187
\\
\midrule
Hierarchical RL & 0.0$\pm$0.0 & 0.662$\pm$0.196 & 0.0$\pm$0.0 & 0.933$\pm$0.021 & 0.0$\pm$0.0 & 0.459$\pm$0.019 & 0.0$\pm$0.0 & 0.918$\pm$0.044\\
RoboTool & 0.207$\pm$ 0.013 & 0.233$\pm$0.003 & 0.033$\pm$0.005 & 0.455$\pm$0.032
& 0.007$\pm$0.005 & 0.627$\pm$0.009 & 0.013$\pm$0.009 & 0.441$\pm$0.002\\
\bottomrule
\end{tabular}

\end{center}
\vspace{-1.5em}
\end{table*}

\subsubsection{Benchmark tasks} We experiment with four benchmark tasks that require strategic high-level reasoning. 
(a) \textbf{Light switching} requires turning off the light with the button high beyond the reach of the robot. The robot needs to determine whether to build a step from boxes based on the height of the stairs, then climb on top of the stairs and stand up to press the button.
(b) \textbf{Package delivery} is delivering a package into a room with the door closed. The robot should ring the doorbell to ask a human inside the room to open the door, and then push the package. 
(c) \textbf{Bridge building} requires the robot to build a bridge to reach the target platform.
(d) \textbf{Elevator ride}, the robot needs to learn how to press the button to call the elevator, and select the correct floor button inside to reach the designated floor.
This task provides the current and target floors, explicitly requiring the robot to decide whether to press the up or down button.
We generate robot code as illustrated in Fig.~\ref{fig:overview} for each task then execute the code in simulation or on the real robot.

\subsubsection{Evaluation metrics}The performances are measured using (i) the overall \textbf{Success} rate for completing the whole task (examples of the desired strategies shown in Fig.~\ref{fig:expr:llm_exp_and_sim}), and (ii) a \textbf{Norm}alized \textbf{Dist}ance metric defined specific to the semantic of each task.
The normalized distance of light switching and elevator ride is calculated as the shortest distance between the robot's toe and the button.
We also report the shortest distance between the package and a fixed spot behind the door for package delivery, as well as the robot's distance from the destination for bridge building. All the distance metrics are normalized by the initial distance.

We use a temperature of 0.2 to generate 3 runs of code for LLM-based methods and evaluate each code for 100 trajectories in simulation.
In each trajectory, we randomize the sizes, initial poses, and physical properties of objects, as well as the dynamics of the robot joints. 
Since RoboTool is designed to generate code for specific parameterized scenarios, we regenerated the code for each trajectory.

\subsection{Ablation study analysis}
\label{sec:Ablation Study Analysis}
As reported in Table~\ref{tab:llm-ablation}, LLM (S+P+C+R) achieves the best performance across tasks. Compared to LLM (S+P+C), the closed-loop planning control provided by the replanner enables the quadruped robot to replan and retry based on error information after failing a challenging subtask (e.g., button pressing), significantly improving the overall task execution success rate. In the light task, if the robot fails to directly climb the lower stair as planned, it recognizes the system's estimation error and replans to decisively push a box as a step, ensuring a more stable and reliable approach to success. However, for the NormDist metric, the closed-loop LLM (S+P+C+R) may retry early steps that do not lower the metric, which leads to a suboptimal score. Removing the parameter calculator and the semantic planner leads to a significant performance drop. 

Note that LLM (S+C) and LLM (C) fail to achieve any success in most tasks since they mostly get stuck at the first executed skill with incorrect arguments. 
LLM (S+C) struggles to correctly consider the bounding box of objects and their geometric relationship.
As shown in Fig.~\ref{fig:expr:llm_exp_and_sim}, the number of spatial miscalculations increases dramatically after the calculator is removed. 
For a more specific example, it sets the target position of climbing to the center of a stair rather than the top surface, and does not consider the robot's height. It also decides to push the movable box exactly beneath the lower stair, which is physically infeasible.
Without a semantic planner for preliminary analysis and planning, LLM (C) is more likely to make incorrect inferences and may even violate constraints, leading to ineffective or dangerous actions. In the bridge building task, it generated an incorrect execution sequence by attempting to push the box after climbing the low platform.

\begin{figure}[h]
    \centering
    \includegraphics[width=\linewidth]{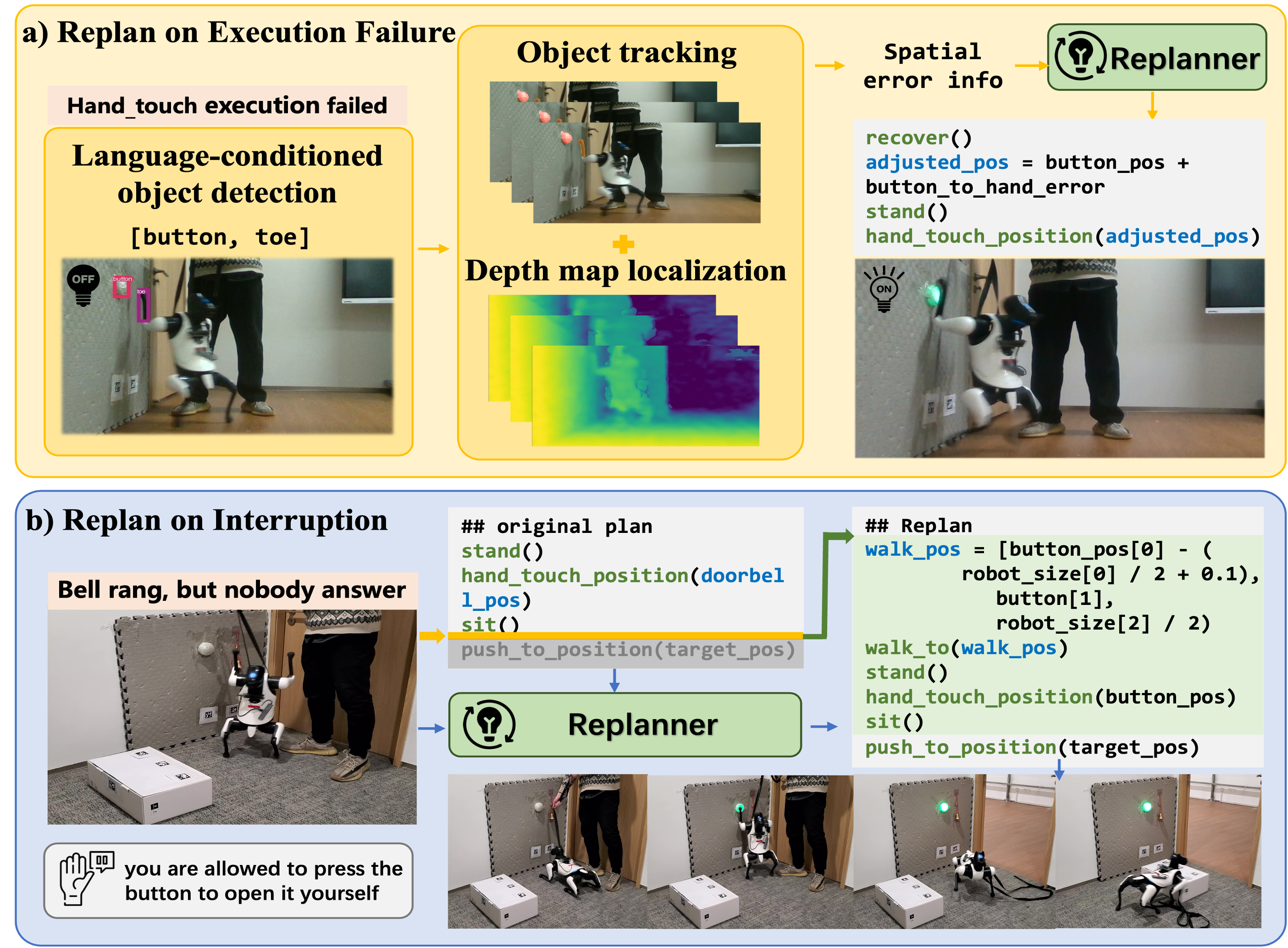}
    \captionsetup{font={small}}
    \caption{Replan deployment. a) Replan on execution failure: After a strategy fails, the robot uses a language-conditioned error detection pipeline to gather spatial information and adjust the parameters. b) Replan on interruption: When the human is unable to open the door but allows the robot to operate the door switch, the robot interrupts the original plan and presses the button to open the door, generating a new code plan while still achieving the long-term delivery goal.}
    \label{fig:expr:replan}
    \vspace{-1.em}
\end{figure}

\begin{figure*}[h]
    \centering
        \includegraphics[width=\linewidth]{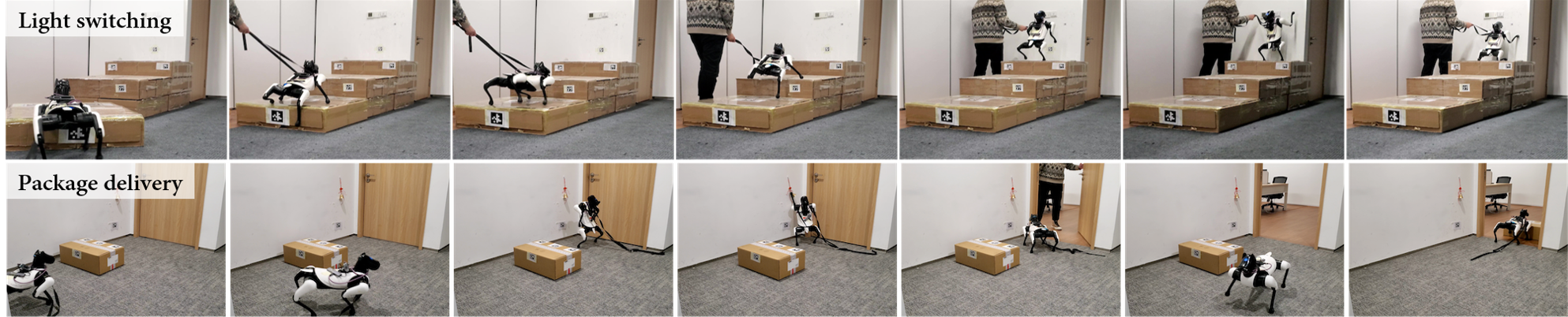}

    \captionsetup{font={small}}
    \caption{Execution of LLM-generated long-term strategies for two benchmark tasks in the real world.}
    \label{fig:expr:real-policy}
    \vspace{-2.5mm}
\end{figure*}

\textbf{Comparison with RoboTool:}
According to Table~\ref{tab:llm-ablation}, RoboTool demonstrates subpar performance across most tasks, which we attribute to three main reasons. 
The code generated by RoboTool lacks branching structures, 
limiting its adaptability to complex tasks.
For instance, elevator ride requires determining whether to press the up or down button based on the current and target floors. Faced with a variety of possible combinations, RoboTool regenerates code for each new scenario, leading to inefficiency and instability. In contrast, our generated code uses an elegant branching structure that covers all possible scenarios for this task.
RoboTool is more likely to make mistakes in spatial calculation since it requires numerical values in the scene object description while simultaneously using variables in the prompts.
As shown in Fig.~\ref{fig:expr:llm_exp_and_sim}, this mixed usage often introduced instability, leading to numerous errors in spatial relationship calculations. 
Furthermore, RoboTool is an open-loop planning framework without a replanner, making it incapable of adjusting its strategy based on failure feedback. 
Relevant prompts and code can be found from our website.

\textbf{Comparison with hierarchical RL:} 
As shown in Table~\ref{tab:llm-ablation}, HRL cannot outperform LLM-based methods. 
For example, in the package delivery task, 
HRL only learns to move the package slightly closer to the door but fails to unlock it.
They both fail to learn the correct solutions since the vast search space makes it almost impossible to explore the exact combination of skills without prior knowledge.

\begin{table}[h]
\captionsetup{font={small}}
    \caption{The effectiveness of fine-tuning low-level RL policies from the terminal states of preceding policies.}
    \label{tab:expr:ablate-finetune}
    \centering
    \begin{tabular}{lcc}
    \toprule
         & Success rate $\uparrow$ & NormDist $\downarrow$\\
    \midrule
        w/ fine-tune & \textbf{0.747$\pm$0.012} & \textbf{0.032$\pm$0.004} \\
        w/o fine-tune & 0.540$\pm$0.020 & 0.177$\pm$0.012\\
    \bottomrule
   \end{tabular}
   \vspace{-1.5em}
\end{table}

\textbf{Ablation on fine-tuning RL skills for robust chaining:}
We compare with a variant without fine-tuning the low-level skills from augmented initial states in the light switching task. As reported in Table~\ref{tab:expr:ablate-finetune}, our method with additional fine-tuning performs significantly better. Since the two methods use the same high-level plans, it indicates that the proposed fine-tuning stage is critical to the robustness of the system when transiting between different skills.

\subsection{Real world experiments}

To facilitate the successful deployment of RL policies to a real Xiaomi Cyberdog2~\cite{cyberdog2} robot, we add domain randomization in motion control and observation noise during training in simulation. 
 We attach multiple AprilTags~\cite{olson2011deploy} to objects and use a forward-view RealSense D430 camera on the robot for pose estimation.
 We also randomly freeze the observations to mimic the situation when the detection fails due to the motion blur of captured images and the loss of target visibility.

\textbf{Replanner Deployment:} 
To support the real-world deployment of replan-on-failure with the necessary failure-related data, we developed a pipeline using off-the-shelf models, as shown in Fig.~\ref{fig:expr:replan}.
A third-person-view camera records the robot's policy execution. Upon failure, the robot identifies key targets for query based on the failed skill. The object detector~\cite{liu2023grounding} guided by the corresponding prompt predicts target bounding boxes, which are then fed into the SAM2 tracker~\cite{ravi2024sam2} to capture the targets' spatial trajectories during execution. Finally, we derive the geometric trajectory by comparing depth maps. For the hand touching policy, this trajectory enables the determination of the minimum distance between the robot’s toe and the button, allowing the replanner to use this offset to refine the code plan.

\textbf{Long-horizon Task:} We test our system on two aforementioned benchmark tasks in the real world. As shown in the upper part of Fig.~\ref{fig:expr:real-policy}, we execute the generated code together with the RL motion skills to turn off the light. In the initial setup, the height of the light button is out of reach with any single skill, and the first level of stairs is too high for the robot to climb directly. 
Considering the scene configuration, the LLM planner decides to push a lower box in front of the existing stairs so that the robot can use it as an intermediate platform to climb up then to touch the button.  
Finally, it runs a sit-down policy to transit to the quadrupedal pose. 

We also show how the quadrupedal robot delivers a package into a room with the door closed initially. To get access to the room, our reasoning layer figures out a strategy to ring a bell to notify the human inside the room to open the door. To protect the package from unnecessary collisions, it leverages the obstacle-avoidance locomotion skill when moving to the bell and before pushing the package.
We test replan-on-interruption as shown in Fig.~\ref{fig:expr:replan}. During the execution of the delivery task, after successfully pressing the doorbell with no response, we manually interrupt the robot’s policy execution and allow it to press the button to open the door. The robot is able to replan based on the new conditions, integrating the previous plan and generating new code to complete the task.

\textbf{Limitations:} 
Currently, the reasoning layer operates over a fixed set of low-level robot skills crafted by experts. Although the system employs a closed-loop structure with the replanner to handle unexpected events, it remains limited to relatively simple scenarios and has yet to be thoroughly evaluated in highly complex or unstructured failure cases.

\section{Conclusion}
In summary, we present an LLM-based system for tackling long-horizon tasks with a quadruped robot that extends the autonomy of quadrupeds from shorter-term motions for specific skills to long-range complex behaviors that require orchestrating multiple locomotion and manipulation policies. The high-level task planning is enabled with our design of multiple LLM agents for decomposing the whole task into robot actions and then grounded to parameterized robot API calls. The low-level control that exploits the agility of quadrupeds to unlock rich environment interactions is instantiated with model-free RL. Building on this work, further research could extend this closed-loop approach to handle more complex tasks and discover new skills autonomously when confronted with unprecedented circumstances.

\section*{Acknowledgements} 
Yi Wu is supported by 2030 Innovation Megaprojects of China(Programme on New Generation Artificial Intelligence) Grant No. 2021AAA0150000.
Z. L. and K. S. acknowledge financical support from InnoHK of the Government of the Hong Kong Special Administrative Region via the Hong Kong Centre for Logistics Robotics. The authors thank Xiaomi Inc. for providing CyberDog2 for experiments.
The authors are grateful to Mengdi Xu for the insightful discussions that significantly contributed to this work.
\noindent

\bibliographystyle{IEEEtran}
\bibliography{reference}

\begin{thebibliography}{10}
\providecommand{\url}[1]{#1}
\csname url@samestyle\endcsname
\providecommand{\newblock}{\relax}
\providecommand{\bibinfo}[2]{#2}
\providecommand{\BIBentrySTDinterwordspacing}{\spaceskip=0pt\relax}
\providecommand{\BIBentryALTinterwordstretchfactor}{4}
\providecommand{\BIBentryALTinterwordspacing}{\spaceskip=\fontdimen2\font plus
\BIBentryALTinterwordstretchfactor\fontdimen3\font minus \fontdimen4\font\relax}
\providecommand{\BIBforeignlanguage}[2]{{%
\expandafter\ifx\csname l@#1\endcsname\relax
\typeout{** WARNING: IEEEtran.bst: No hyphenation pattern has been}%
\typeout{** loaded for the language `#1'. Using the pattern for}%
\typeout{** the default language instead.}%
\else
\language=\csname l@#1\endcsname
\fi
#2}}
\providecommand{\BIBdecl}{\relax}
\BIBdecl

\bibitem{de2006quadrupedal}
P.~G. De~Santos, E.~Garcia, and J.~Estremera, \emph{Quadrupedal locomotion: an introduction to the control of four-legged robots}.\hskip 1em plus 0.5em minus 0.4em\relax Springer, 2006, vol.~1.

\bibitem{fankhauser2018robust}
P.~Fankhauser, M.~Bjelonic, C.~D. Bellicoso, T.~Miki, and M.~Hutter, ``Robust rough-terrain locomotion with a quadrupedal robot,'' in \emph{2018 IEEE International Conference on Robotics and Automation (ICRA)}.\hskip 1em plus 0.5em minus 0.4em\relax IEEE, 2018, pp. 5761--5768.

\bibitem{lee2020learning}
J.~Lee, J.~Hwangbo, L.~Wellhausen, V.~Koltun, and M.~Hutter, ``Learning quadrupedal locomotion over challenging terrain,'' \emph{Science robotics}, vol.~5, no.~47, p. eabc5986, 2020.

\bibitem{shi2021circus}
F.~Shi, T.~Homberger, J.~Lee, T.~Miki, M.~Zhao, F.~Farshidian, K.~Okada, M.~Inaba, and M.~Hutter, ``Circus anymal: A quadruped learning dexterous manipulation with its limbs,'' in \emph{2021 IEEE International Conference on Robotics and Automation (ICRA)}.\hskip 1em plus 0.5em minus 0.4em\relax IEEE, 2021, pp. 2316--2323.

\bibitem{sombolestan2023hierarchical}
M.~Sombolestan and Q.~Nguyen, ``Hierarchical adaptive loco-manipulation control for quadruped robots,'' in \emph{2023 IEEE International Conference on Robotics and Automation (ICRA)}.\hskip 1em plus 0.5em minus 0.4em\relax IEEE, 2023, pp. 12\,156--12\,162.

\bibitem{arm2024pedipulate}
P.~Arm, M.~Mittal, H.~Kolvenbach, and M.~Hutter, ``Pedipulate: Enabling manipulation skills using a quadruped robot's leg,'' 2024.

\bibitem{cheng23legs}
X.~Cheng, A.~Kumar, and D.~Pathak, ``Legs as manipulator: Pushing quadrupedal agility beyond locomotion,'' in \emph{{IEEE} International Conference on Robotics and Automation, {ICRA} 2023, London, UK, May 29 - June 2, 2023}.\hskip 1em plus 0.5em minus 0.4em\relax {IEEE}, 2023, pp. 5106--5112.

\bibitem{garrett2015ffrob}
C.~R. Garrett, T.~Lozano-P{\'e}rez, and L.~P. Kaelbling, ``Ffrob: An efficient heuristic for task and motion planning,'' in \emph{Algorithmic Foundations of Robotics XI: Selected Contributions of the Eleventh International Workshop on the Algorithmic Foundations of Robotics}.\hskip 1em plus 0.5em minus 0.4em\relax Springer, 2015, pp. 179--195.

\bibitem{hwangbo2019learning}
J.~Hwangbo, J.~Lee, A.~Dosovitskiy, D.~Bellicoso, V.~Tsounis, V.~Koltun, and M.~Hutter, ``Learning agile and dynamic motor skills for legged robots,'' \emph{Science Robotics}, vol.~4, no.~26, p. eaau5872, 2019.

\bibitem{miki2022learning}
T.~Miki, J.~Lee, J.~Hwangbo, L.~Wellhausen, V.~Koltun, and M.~Hutter, ``Learning robust perceptive locomotion for quadrupedal robots in the wild,'' \emph{Science Robotics}, vol.~7, no.~62, p. eabk2822, 2022.

\bibitem{zhuang2023robot}
Z.~Zhuang, Z.~Fu, J.~Wang, C.~G. Atkeson, S.~Schwertfeger, C.~Finn, and H.~Zhao, ``Robot parkour learning,'' in \emph{Conference on Robot Learning}.\hskip 1em plus 0.5em minus 0.4em\relax PMLR, 2023, pp. 73--92.

\bibitem{cheng2023extreme}
X.~Cheng, K.~Shi, A.~Agarwal, and D.~Pathak, ``Extreme parkour with legged robots,'' \emph{arXiv preprint arXiv:2309.14341}, 2023.

\bibitem{jeon2023learning}
S.~Jeon, M.~Jung, S.~Choi, B.~Kim, and J.~Hwangbo, ``Learning whole-body manipulation for quadrupedal robot,'' \emph{IEEE Robotics and Automation Letters}, vol.~9, no.~1, pp. 699--706, 2023.

\bibitem{schwarke2023curiosity}
C.~Schwarke, V.~Klemm, M.~Van~der Boon, M.~Bjelonic, and M.~Hutter, ``Curiosity-driven learning of joint locomotion and manipulation tasks,'' in \emph{Proceedings of The 7th Conference on Robot Learning}, vol. 229.\hskip 1em plus 0.5em minus 0.4em\relax PMLR, 2023, pp. 2594--2610.

\bibitem{bellicoso2019alma}
C.~D. Bellicoso, K.~Kr{\"a}mer, M.~St{\"a}uble, D.~Sako, F.~Jenelten, M.~Bjelonic, and M.~Hutter, ``Alma-articulated locomotion and manipulation for a torque-controllable robot,'' in \emph{2019 International conference on robotics and automation (ICRA)}.\hskip 1em plus 0.5em minus 0.4em\relax IEEE, 2019, pp. 8477--8483.

\bibitem{fu2023deep}
Z.~Fu, X.~Cheng, and D.~Pathak, ``Deep whole-body control: learning a unified policy for manipulation and locomotion,'' in \emph{Conference on Robot Learning}.\hskip 1em plus 0.5em minus 0.4em\relax PMLR, 2023, pp. 138--149.

\bibitem{ji2022hierarchical}
Y.~Ji, Z.~Li, Y.~Sun, X.~B. Peng, S.~Levine, G.~Berseth, and K.~Sreenath, ``Hierarchical reinforcement learning for precise soccer shooting skills using a quadrupedal robot,'' in \emph{2022 IEEE/RSJ International Conference on Intelligent Robots and Systems (IROS)}.\hskip 1em plus 0.5em minus 0.4em\relax IEEE, 2022, pp. 1479--1486.

\bibitem{huang2023creating}
X.~Huang, Z.~Li, Y.~Xiang, Y.~Ni, Y.~Chi, Y.~Li, L.~Yang, X.~B. Peng, and K.~Sreenath, ``Creating a dynamic quadrupedal robotic goalkeeper with reinforcement learning,'' in \emph{2023 IEEE/RSJ International Conference on Intelligent Robots and Systems (IROS)}.\hskip 1em plus 0.5em minus 0.4em\relax IEEE, 2023, pp. 2715--2722.

\bibitem{garrett2021integrated}
C.~R. Garrett, R.~Chitnis, R.~Holladay, B.~Kim, T.~Silver, L.~P. Kaelbling, and T.~Lozano-P{\'e}rez, ``Integrated task and motion planning,'' \emph{Annual review of control, robotics, and autonomous systems}, vol.~4, pp. 265--293, 2021.

\bibitem{hartmann2022long}
V.~N. Hartmann, A.~Orthey, D.~Driess, O.~S. Oguz, and M.~Toussaint, ``Long-horizon multi-robot rearrangement planning for construction assembly,'' \emph{IEEE Transactions on Robotics}, vol.~39, no.~1, pp. 239--252, 2022.

\bibitem{gupta2020relay}
A.~Gupta, V.~Kumar, C.~Lynch, S.~Levine, and K.~Hausman, ``Relay policy learning: Solving long-horizon tasks via imitation and reinforcement learning,'' in \emph{Conference on Robot Learning}.\hskip 1em plus 0.5em minus 0.4em\relax PMLR, 2020, pp. 1025--1037.

\bibitem{yang2021hierarchical}
X.~Yang, Z.~Ji, J.~Wu, Y.-K. Lai, C.~Wei, G.~Liu, and R.~Setchi, ``Hierarchical reinforcement learning with universal policies for multistep robotic manipulation,'' \emph{IEEE Transactions on Neural Networks and Learning Systems}, vol.~33, no.~9, pp. 4727--4741, 2021.

\bibitem{zhu2022bottom}
Y.~Zhu, P.~Stone, and Y.~Zhu, ``Bottom-up skill discovery from unsegmented demonstrations for long-horizon robot manipulation,'' \emph{IEEE Robotics and Automation Letters}, vol.~7, no.~2, pp. 4126--4133, 2022.

\bibitem{lee2022adversarial}
Y.~Lee, J.~J. Lim, A.~Anandkumar, and Y.~Zhu, ``Adversarial skill chaining for long-horizon robot manipulation via terminal state regularization,'' in \emph{Conference on Robot Learning}.\hskip 1em plus 0.5em minus 0.4em\relax PMLR, 2022, pp. 406--416.

\bibitem{brown2020language}
T.~Brown, B.~Mann, N.~Ryder, M.~Subbiah, J.~D. Kaplan, P.~Dhariwal, A.~Neelakantan, P.~Shyam, G.~Sastry, A.~Askell \emph{et~al.}, ``Language models are few-shot learners,'' \emph{Advances in neural information processing systems}, vol.~33, pp. 1877--1901, 2020.

\bibitem{ouyang2022training}
L.~Ouyang, J.~Wu, X.~Jiang, D.~Almeida, C.~Wainwright, P.~Mishkin, C.~Zhang, S.~Agarwal, K.~Slama, A.~Ray \emph{et~al.}, ``Training language models to follow instructions with human feedback,'' \emph{Advances in Neural Information Processing Systems}, vol.~35, pp. 27\,730--27\,744, 2022.

\bibitem{chowdhery2023palm}
A.~Chowdhery, S.~Narang, J.~Devlin, M.~Bosma, G.~Mishra, A.~Roberts, P.~Barham, H.~W. Chung, C.~Sutton, S.~Gehrmann \emph{et~al.}, ``Palm: Scaling language modeling with pathways,'' \emph{Journal of Machine Learning Research}, vol.~24, no. 240, pp. 1--113, 2023.

\bibitem{brohan2023can}
A.~Brohan, Y.~Chebotar, C.~Finn, K.~Hausman, A.~Herzog, D.~Ho, J.~Ibarz, A.~Irpan, E.~Jang, R.~Julian \emph{et~al.}, ``Do as i can, not as i say: Grounding language in robotic affordances,'' in \emph{Conference on Robot Learning}.\hskip 1em plus 0.5em minus 0.4em\relax PMLR, 2023, pp. 287--318.

\bibitem{huang2022language}
W.~Huang, P.~Abbeel, D.~Pathak, and I.~Mordatch, ``Language models as zero-shot planners: Extracting actionable knowledge for embodied agents,'' in \emph{International Conference on Machine Learning}.\hskip 1em plus 0.5em minus 0.4em\relax PMLR, 2022, pp. 9118--9147.

\bibitem{huang2023inner}
W.~Huang, F.~Xia, T.~Xiao, H.~Chan, J.~Liang, P.~Florence, A.~Zeng, J.~Tompson, I.~Mordatch, Y.~Chebotar \emph{et~al.}, ``Inner monologue: Embodied reasoning through planning with language models,'' in \emph{Conference on Robot Learning}.\hskip 1em plus 0.5em minus 0.4em\relax PMLR, 2023, pp. 1769--1782.

\bibitem{zeng2022socratic}
A.~Zeng, M.~Attarian, K.~M. Choromanski, A.~Wong, S.~Welker, F.~Tombari, A.~Purohit, M.~S. Ryoo, V.~Sindhwani, J.~Lee \emph{et~al.}, ``Socratic models: Composing zero-shot multimodal reasoning with language,'' in \emph{The Eleventh International Conference on Learning Representations}, 2022.

\bibitem{liang2023code}
J.~Liang, W.~Huang, F.~Xia, P.~Xu, K.~Hausman, B.~Ichter, P.~Florence, and A.~Zeng, ``Code as policies: Language model programs for embodied control,'' in \emph{2023 IEEE International Conference on Robotics and Automation (ICRA)}.\hskip 1em plus 0.5em minus 0.4em\relax IEEE, 2023, pp. 9493--9500.

\bibitem{xu2023creative}
M.~Xu, W.~Yu, P.~Huang, S.~Liu, X.~Zhang, Y.~Niu, T.~Zhang, F.~Xia, J.~Tan, and D.~Zhao, ``Creative robot tool use with large language models,'' in \emph{NeurIPS 2023 Foundation Models for Decision Making Workshop}, 2023.

\bibitem{achiam2023gpt}
J.~Achiam, S.~Adler, S.~Agarwal, L.~Ahmad, I.~Akkaya, F.~L. Aleman, D.~Almeida, J.~Altenschmidt, S.~Altman, S.~Anadkat \emph{et~al.}, ``Gpt-4 technical report,'' \emph{arXiv preprint arXiv:2303.08774}, 2023.

\bibitem{yu2023language}
W.~Yu, N.~Gileadi, C.~Fu, S.~Kirmani, K.-H. Lee, M.~G. Arenas, H.-T.~L. Chiang, T.~Erez, L.~Hasenclever, J.~Humplik \emph{et~al.}, ``Language to rewards for robotic skill synthesis,'' \emph{arXiv preprint arXiv:2306.08647}, 2023.

\bibitem{schulman2017proximal}
J.~Schulman, F.~Wolski, P.~Dhariwal, A.~Radford, and O.~Klimov, ``Proximal policy optimization algorithms,'' \emph{arXiv preprint arXiv:1707.06347}, 2017.

\bibitem{margolis2023walk}
G.~B. Margolis and P.~Agrawal, ``Walk these ways: Tuning robot control for generalization with multiplicity of behavior,'' in \emph{Conference on Robot Learning}.\hskip 1em plus 0.5em minus 0.4em\relax PMLR, 2023, pp. 22--31.

\bibitem{li2023learning}
Y.~Li, J.~Li, W.~Fu, and Y.~Wu, ``Learning agile bipedal motions on a quadrupedal robot,'' \emph{arXiv preprint arXiv:2311.05818}, 2023.

\bibitem{cyberdog2}
Xiaomi, ``Cyberdog2,'' \url{https://www.mi.com/cyberdog2}, 2023, accessed: Mar. 2024.

\bibitem{olson2011deploy}
E.~Olson, ``Apriltag: A robust and flexible visual fiducial system,'' in \emph{2011 IEEE International Conference on Robotics and Automation}, 2011, pp. 3400--3407.

\bibitem{liu2023grounding}
S.~Liu, Z.~Zeng, T.~Ren, F.~Li, H.~Zhang, J.~Yang, C.~Li, J.~Yang, H.~Su, J.~Zhu \emph{et~al.}, ``Grounding dino: Marrying dino with grounded pre-training for open-set object detection,'' \emph{arXiv preprint arXiv:2303.05499}, 2023.

\bibitem{ravi2024sam2}
N.~Ravi, V.~Gabeur, Y.-T. Hu, R.~Hu, C.~Ryali, T.~Ma, H.~Khedr, R.~R{\"a}dle, C.~Rolland, L.~Gustafson, E.~Mintun, J.~Pan, K.~V. Alwala, N.~Carion, C.-Y. Wu, R.~Girshick, P.~Doll{\'a}r, and C.~Feichtenhofer, ``Sam 2: Segment anything in images and videos,'' \emph{arXiv preprint arXiv:2408.00714}, 2024.

\end{thebibliography}


\begin{thebibliography}{99}

\bibitem{c1} G. O. Young, ÒSynthetic structure of industrial plastics (Book style with paper title and editor),Ó 	in Plastics, 2nd ed. vol. 3, J. Peters, Ed.  New York: McGraw-Hill, 1964, pp. 15Ð64.
\bibitem{c2} W.-K. Chen, Linear Networks and Systems (Book style).	Belmont, CA: Wadsworth, 1993, pp. 123Ð135.
\bibitem{c3} H. Poor, An Introduction to Signal Detection and Estimation.   New York: Springer-Verlag, 1985, ch. 4.
\bibitem{c4} B. Smith, ÒAn approach to graphs of linear forms (Unpublished work style),Ó unpublished.
\bibitem{c5} E. H. Miller, ÒA note on reflector arrays (Periodical styleÑAccepted for publication),Ó IEEE Trans. Antennas Propagat., to be publised.
\bibitem{c6} J. Wang, ÒFundamentals of erbium-doped fiber amplifiers arrays (Periodical styleÑSubmitted for publication),Ó IEEE J. Quantum Electron., submitted for publication.
\bibitem{c7} C. J. Kaufman, Rocky Mountain Research Lab., Boulder, CO, private communication, May 1995.
\bibitem{c8} Y. Yorozu, M. Hirano, K. Oka, and Y. Tagawa, ÒElectron spectroscopy studies on magneto-optical media and plastic substrate interfaces(Translation Journals style),Ó IEEE Transl. J. Magn.Jpn., vol. 2, Aug. 1987, pp. 740Ð741 [Dig. 9th Annu. Conf. Magnetics Japan, 1982, p. 301].
\bibitem{c9} M. Young, The Techincal Writers Handbook.  Mill Valley, CA: University Science, 1989.
\bibitem{c10} J. U. Duncombe, ÒInfrared navigationÑPart I: An assessment of feasibility (Periodical style),Ó IEEE Trans. Electron Devices, vol. ED-11, pp. 34Ð39, Jan. 1959.
\bibitem{c11} S. Chen, B. Mulgrew, and P. M. Grant, ÒA clustering technique for digital communications channel equalization using radial basis function networks,Ó IEEE Trans. Neural Networks, vol. 4, pp. 570Ð578, July 1993.
\bibitem{c12} R. W. Lucky, ÒAutomatic equalization for digital communication,Ó Bell Syst. Tech. J., vol. 44, no. 4, pp. 547Ð588, Apr. 1965.
\bibitem{c13} S. P. Bingulac, ÒOn the compatibility of adaptive controllers (Published Conference Proceedings style),Ó in Proc. 4th Annu. Allerton Conf. Circuits and Systems Theory, New York, 1994, pp. 8Ð16.
\bibitem{c14} G. R. Faulhaber, ÒDesign of service systems with priority reservation,Ó in Conf. Rec. 1995 IEEE Int. Conf. Communications, pp. 3Ð8.
\bibitem{c15} W. D. Doyle, ÒMagnetization reversal in films with biaxial anisotropy,Ó in 1987 Proc. INTERMAG Conf., pp. 2.2-1Ð2.2-6.
\bibitem{c16} G. W. Juette and L. E. Zeffanella, ÒRadio noise currents n short sections on bundle conductors (Presented Conference Paper style),Ó presented at the IEEE Summer power Meeting, Dallas, TX, June 22Ð27, 1990, Paper 90 SM 690-0 PWRS.
\bibitem{c17} J. G. Kreifeldt, ÒAn analysis of surface-detected EMG as an amplitude-modulated noise,Ó presented at the 1989 Int. Conf. Medicine and Biological Engineering, Chicago, IL.
\bibitem{c18} J. Williams, ÒNarrow-band analyzer (Thesis or Dissertation style),Ó Ph.D. dissertation, Dept. Elect. Eng., Harvard Univ., Cambridge, MA, 1993. 
\bibitem{c19} N. Kawasaki, ÒParametric study of thermal and chemical nonequilibrium nozzle flow,Ó M.S. thesis, Dept. Electron. Eng., Osaka Univ., Osaka, Japan, 1993.
\bibitem{c20} J. P. Wilkinson, ÒNonlinear resonant circuit devices (Patent style),Ó U.S. Patent 3 624 12, July 16, 1990. 






\end{thebibliography}

\end{document}